\documentclass{isprs} 
\usepackage{subfigure}
\usepackage{setspace}
\usepackage{geometry} 
\usepackage{epstopdf}
\usepackage[labelsep=period]{caption}  
\usepackage[british]{babel} 
\usepackage[hang]{footmisc}

\usepackage{makecell}
\usepackage{hyperref}
\usepackage{dblfloatfix}
\usepackage[fleqn]{amsmath}
\usepackage{todonotes}

\begin{document}

\title{Analyzing the impact of semantic LoD3 building models on image-based vehicle localization}
\date{}



\author{Antonia Bieringer\textsuperscript{1}, Olaf Wysocki\textsuperscript{1}, Sebastian Tuttas\textsuperscript{3}, Ludwig Hoegner\textsuperscript{4}, Christoph Holst\textsuperscript{1,2}}

\address{


	\textsuperscript{1 }Professorship of Photogrammetry and Remote Sensing, 
    \textsuperscript{2 }Chair of Engineering Geodesy, \\
 TUM School of Engineering and Design, Technical University of Munich, 80333 Munich, Germany \\ - (antonia.bieringer, olaf.wysocki, christoph.holst)@tum.de\\
 \textsuperscript{3 }3D Mapping Solutions GmbH, 83607 Holzkirchen, Germany - sebastian.tuttas@3d-mapping.de\\
\textsuperscript{4 }Hochschule München University of Applied Sciences, 80335 Munich, Germany - ludwig.hoegner@hm.edu\\

}



\abstract{

Numerous navigation applications rely on data from global navigation satellite systems (GNSS), even though their accuracy is compromised in urban areas, posing a significant challenge, particularly for precise autonomous car localization.
Extensive research has focused on enhancing localization accuracy by integrating various sensor types to address this issue.
This paper introduces a novel approach for car localization, leveraging image features that correspond with highly detailed semantic 3D building models.
The core concept involves augmenting positioning accuracy by incorporating prior geometric and semantic knowledge into calculations. 
The work assesses outcomes using Level of Detail 2 (LoD2) and Level of Detail 3 (LoD3) models, analyzing whether facade-enriched models yield superior accuracy. 
This comprehensive analysis encompasses diverse methods, including off-the-shelf feature matching and deep learning, facilitating thorough discussion. 
Our experiments corroborate that LoD3 enables detecting up to 69\% more features than using LoD2 models.
We believe that this study will contribute to the research of enhancing positioning accuracy in GNSS-denied urban canyons. 
It also shows a practical application of under-explored LoD3 building models on map-based car positioning.
}

\keywords{LoD2, LoD3, semantic 3D city models, car localization, map-based localization, image-based positioning}

\maketitle


 
\sloppy

\section{Introduction}
Navigating in complex urban environments can pose challenges for positioning systems, primarily when vehicles depend on GNSS. As vehicles move through signal-obstructed urban canyons, the effectiveness of positioning systems can diminish. 
Consequently, vehicles shall turn to alternative cues to determine their location in such scenarios.

Various methods have been developed to address this issue, combining different sensors to enhance location accuracy. 
While GNSS enables global positioning, supporting it with local cues is an intuitive choice. 
For example, directly enhancing GNSS signals by using 3D city models \cite{dreier2021strategien}.
More frequently, however, the researchers have focused on utilizing cameras for enhanced positioning owing to their abundance on cars and optical features.

In order to localize the vehicle with optical cameras, features have to be found in the images. There are various techniques for feature finding and matching, e.g., SIFT \cite{lowe2004SIFT}, SURF \cite{bay2008SURF}, or ORB \cite{rublee2011ORB}. 
Over time, researchers compared those methods against each other \cite{bansal2021compSSO} and improved them.
However, to globally position the images, a reference 3D map with the correct scale and position shall be used.

Such 3D map can be extracted directly from governmental and proprietary geoportals \cite{lucks2021LOD2}, where semantic 3D building models are available.
These models are used in different Level of Details (LoD). LoD1 models are limited to geometrical information that includes the vertices and edges of the buildings; they are cuboid polyhedral models with flat roofs. 
LoD2 models are more detailed and include the generalized structure of the roof. 
In comparison, LoD3 is enhanced by facade details such as windows and doors \cite{biljecki2016LoD}. 

A few papers have already investigated the usage of LoD1 and LoD2 models for vehicle localization.
\begin{figure*}[!hb] 
    \centering
    \includegraphics[width=\linewidth]{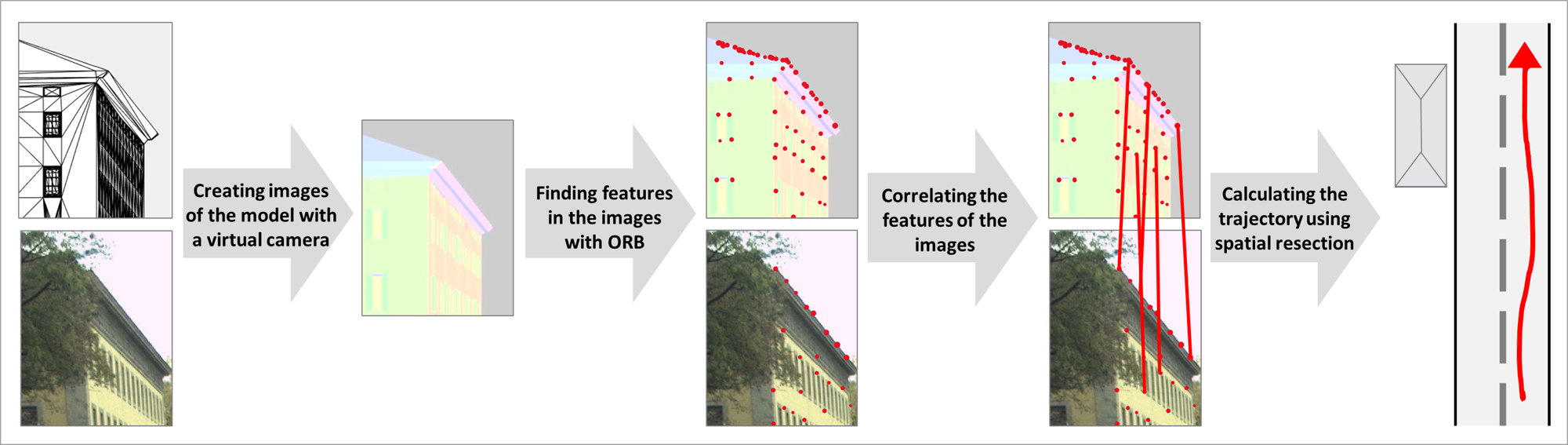}
    \caption{Developed workflow: The images of the real world are matched with virtual images of the LoD models. The combination of both will be used to calculate the trajectory.}
    \label{fig:workflow}
\end{figure*}
One of the examples of utilizing LoD1 and LoD2 models is the paper of Vogel et al., who introduce a localizing approach in the inner-city outdoor environments using 3D city models \cite{vogel2018citymodel}. 
To this end, they modify an iterated extended Kalman filter to be able to integrate additional information into it, e.g., available maps. 
A more reliable and precise georeferencing is achieved since geometric circumstances are considered \cite{vogel2018citymodel}. 
Their idea is refined in later papers of \cite{moftizadeh2021LOD2,lucks2021LOD2}.
However, the usage of LoD3 models facade features for localization has yet to be investigated.

Our contributions are as follows:

\begin{itemize}
 \item Map-supported vehicle localization by using high fidelity LoD3 models and images.
 \item Harmonizing two distinct modalities: Visual information encapsulated within optical images and the structure of semantic 3D models' non-textured surfaces.
 \item Comparing the benefits of LoD3 over LoD2 building models for camera localization.
\end{itemize}

\section{Related Work}
We devote our work to analyzing the impact of image-based vehicle positioning using semantics of 3D maps. 
Therefore, in this Section, we present the state-of-the-art in semantic 3D city models and image- and map-based positioning.
\subsection{Semantic 3D city models}
At city, regional, and national scales, semantic 3D city models offer comprehensive representations of structures, taxonomies, and aggregations.
Semantic 3D city models are frequently described by the internationally adopted CityGML standard, established by the Open Geospatial Consortium (OGC) \cite{grogerOGCCityGeography2012}; 
This data model standard adopts two encodings using either Geography Markup Language (GML) or CityJSON \cite{Kutzner2020,ledoux2019cityjson}.
The CityGML standard enables the modeling of urban objects with 3D geometry, appearance, topology, and semantics across four different Levels of Detail (LoD). 

Building descriptions are pivotal to semantic 3D city models due to the foundational role of urban dwellings in cities \cite{biljeckiApplications3DCity2015}. 
LoD1 and LoD2 building models, currently prevalent, boast around 220 million models available in countries such as Germany, Japan, the Netherlands, Switzerland, the United States, and Poland\footnotemark[1].
While LoD1 building models are represented by prismatic 3D models of height-extruded building footprints, LoD2 models additionally represent complex roof structures.     
The widespread adoption of these models is attributed to robust 3D reconstruction algorithms and the availability of building footprints combined with aerial observations \cite{RoschlaubBatscheider,HAALA2010570}.

However, LoD1 and LoD2 models lack a more detailed description of facade elements.
Addressing this gap are LoD3 building models featuring descriptive facades with objects such as windows, doors, balconies, and underpasses \cite{wysockiUnderpasses}.
The current practice shows that these highly-detailed objects are manually modeled \cite{uggla2023future,manualLoD3seismic}. 
Yet, a great deal of research has been devoted to enabling automatic LoD3 reconstruction, which has sparked advancements in the LoD3 data adoption
\footnote{\url{https://github.com/OloOcki/awesome-citygml}}.
%
The recently developed LoD3 reconstruction methods promise to increase the availability of such models \cite{wysocki2023scan2lod3,hoegner2022automatic,helmutMayerLoD3}.

\subsection{Image-Based Vehicle Positioning}


One way of reconstructing the 3D structure of a scene and estimating the camera poses from a collection of 2D images is Structure from Motion (SfM) \cite{schoenberger2016sfm}. 
This technique works by leveraging the relationships between features in different images and the scene's geometry to recover the scene's 3D structure and camera pose.

Another approach of estimating the vehicle's position by using a series of camera images is Visual Odometry (VO). 
It is advantageous over other methods due to its insensitivity to soil mechanics and lower drift rates \cite{howard2008vo}. 
Additionally, it performs well in places where a satellite connection is limited, e.g., indoors or on narrow streets. 
However, the trajectories estimated by visual odometry tend to drift along with increasing distance from the start point. Consequently, researchers investigated to improve the concept by expanding VO ideas to another image-based localization method called SLAM \cite{gao2019vSLAM}.

Simultaneous Localization and Mapping (SLAM) reconstructs reality from observation data, e.g., images and points clouds. Compared to SfM and VO, it computes the trajectory in real time and maps the vehicle's environment \cite{gao2019vSLAM}. The map is updated while driving \cite{rehder2015submap}, and consequently, the current point is automatically corrected by previously calculated results. 
A prerequisite for the map-supported SLAM are surroundings that are easy to detect and can be used to scale and correctly map the trajectory \cite{hungar2020map}. 
As Lucks et al. show, the prior 3D building map can enhance the SLAM-based calculated trajectory \cite{lucks2021LOD2}. 

\section{Methodology}
In Figure \ref{fig:workflow}, the workflow of the developed method is shown. 
First, virtual pictures of the building models are created. The resulting images are then used to find features in the corresponding optical images. 
For this purpose, the ORB detector \cite{rublee2011ORB} is used. The generated 2D-image coordinates are correlated with a 3D coordinate by benefiting from the LoD model semantics and geometry. In the end, these coordinates are needed to estimate the camera position for every single image with spatial resection. 
This workflow is designed for LoD1, LoD2 and LoD3 building models. 
The developed method's code is published in the GitHub repository\footnote{\url{https://github.com/tum-pf/LoD3ForLocalization}}.
%

\subsection{Creating Virtual Images of the LoD Models}

For the first step, the virtual images are created by using ray casting. To work with the ray casting algorithm, a polyhedral building model is tranformed into a mesh model using triangulation since all results are based on the hit of a triangle by virtual rays. 
The algorithm creates those virtual rays and sends them through a virtual camera, preferably in the direction of the model. Every time a ray hits a point, the corresponding triangle and the barycentric coordinates of the exact hit point in the triangle are saved. Therefore, the 3D coordinates in the real world can be calculated by only knowing the 2D coordinates of the virtual image. The calculation is described in Section \ref{sec:CorrelationWith3DCoord}.

\begin{figure}[h!]
    \centering
    \includegraphics[width=2.5in]{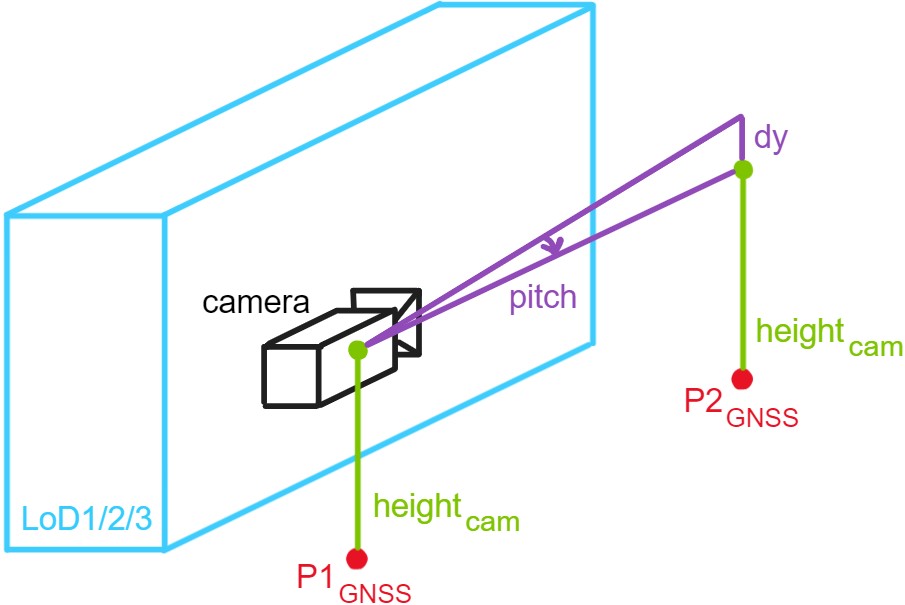}
    \caption{Virtual setup for ray casting.}
    \label{fig:setup_raycasting}
\end{figure}

Moreover, the virtual camera's pose is required (see Figure \ref{fig:setup_raycasting}). As an approximation solution, the position is taken from the GNSS data of the current corresponding optical image. 
On top, the camera's height above the GNSS antenna must be considered, leading to GNSS values adapted to the camera position. 
The camera is pointing to the adapted GNSS position of the consecutive image. 
Afterward, the orientation of the camera requires to be adjusted: The angles for roll, pitch, and yaw are required to approximate the point close to the GNSS position in consideration of the camera tilt. 
A visual example of pitch can be seen in Figure \ref{fig:setup_raycasting}. The approximation is calculated as follows:
\begin{equation}
dy [m] = r_{GNSS} \cdot dy  [^{\circ}] \cdot \frac{\pi} {180 ^{\circ}}
\end{equation}
An overview of the above-described virtual setup can be seen in Figure \ref{fig:setup_raycasting}.

After applying the ray casting algorithm, the results are available in a tensor. There are five results: the distance, the geometry IDs, the primitive IDs, the primitive normals, and the barycentric coordinates. For an example of LoD2 combined with one LoD3 building, see Figure \ref{fig:results_raycastingLoD3}:

\begin{figure}[h]
    \centering
    \includegraphics[width=\linewidth]{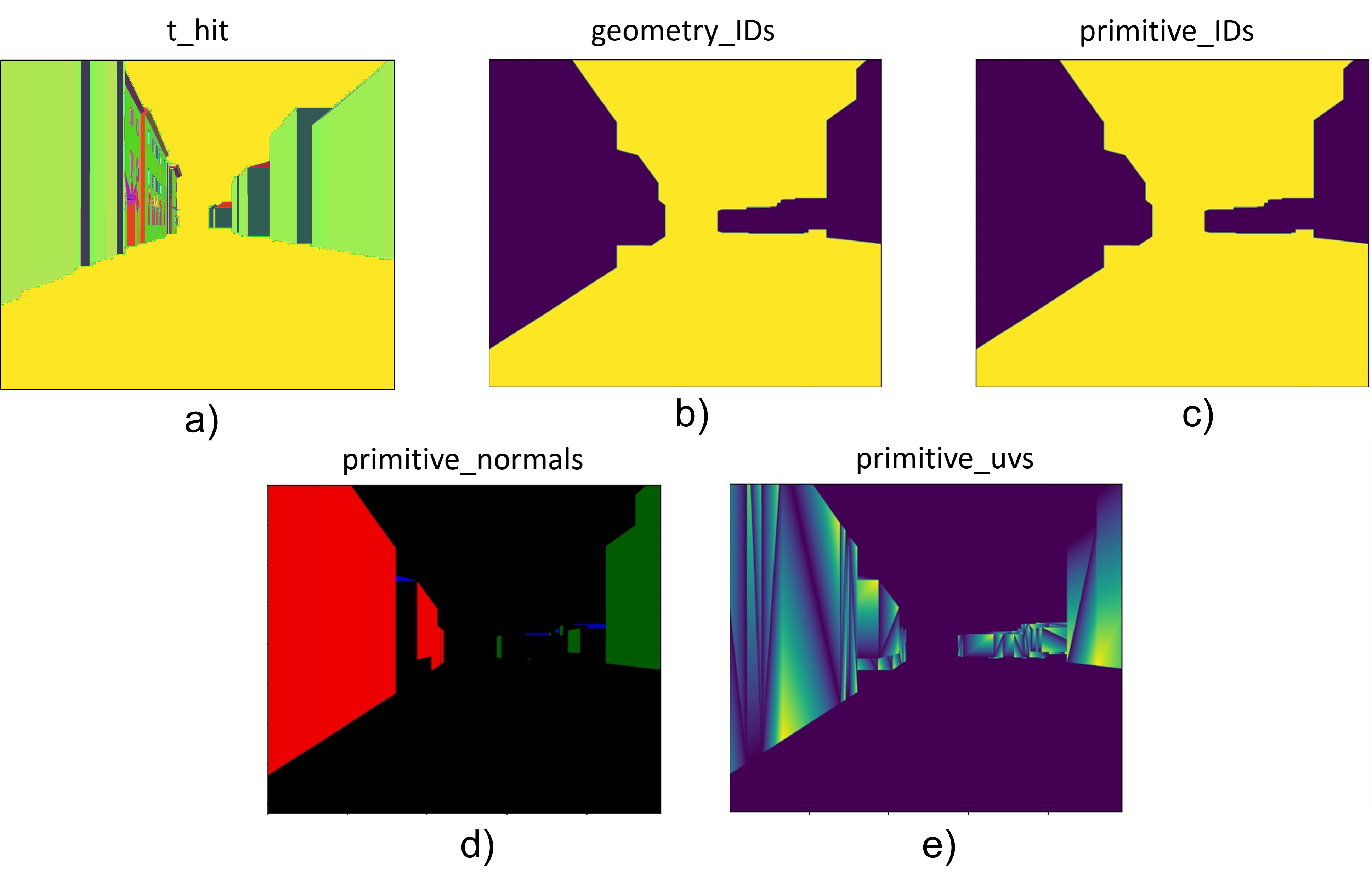}
    \caption{Ray casting results for LoD3 a) hit distance b) geometry IDs c) primitive IDs d) primitive normals e) barycentric coordinates.}
    \label{fig:results_raycastingLoD3}
\end{figure}
From these results, the primitive-normals-data is saved as an image. As shown in Figure \ref{fig:results_raycastingLoD3}d, some walls of the buildings have the same color as the background. To make every plane of the building visible, the values of the primitive normals are adjusted so that only the absolute values are considered since the exact value is unimportant for the virtual images.
This adjustment is necessary so that every side of the building has a unique color. Otherwise, some buildings' planes will not be considered during feature matching. 
Note that the LoD3 building in the virtual image in Figure \ref{fig:results_raycastingLoD3} is more detailed due to the additional information on the model, especially regarding windows and roof eaves.

\subsection{Feature Matching}

\begin{figure}[h!]
    \centering
    \includegraphics[width=3.3in]{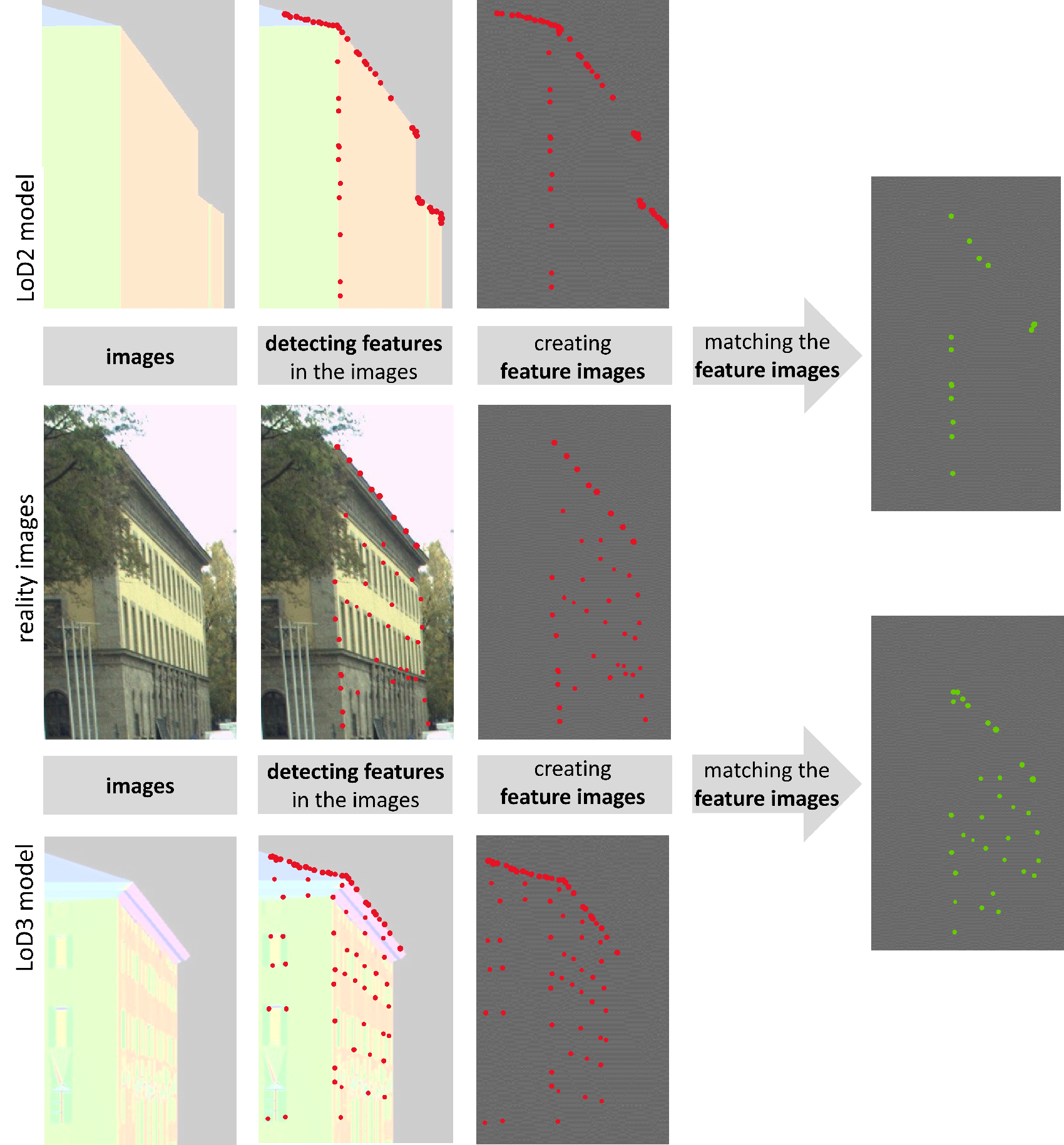}
    \caption{Workflow of the generation of the feature images.}
    \label{fig:FeatureImages}
\end{figure}

An overview of the approach is visualized in Figure \ref{fig:FeatureImages}. The feature matching is performed on images with reduced details. Therefore, the images become more abstract and only focus on important information. 
Then, the images are matched against themselves: the virtual and the optical image. 

The next step is done for the image of the real world, as well as for the one of the virtual camera. 
First, another abstract, virtual feature image is created and implemented as a matrix with the same values for every pixel. If features were found in matching the images against themselves, those pixels are colored differently. This is realized by assigning a different value to this pixel in the matrix. The image, therefore, only contains information about the essential points of the image before. 

Afterward, these newly created feature images are matched with an ORB-feature detector again. 

\subsection{Correlating the Features with a 3D-Coordinate} \label{sec:CorrelationWith3DCoord}

During ray casting, every hit of the mesh by a ray is saved. 
In this case, the corresponding triangle IDs are stored. 
Then, the corresponding triangles in the mesh are found, which enables the extraction of which vertices belong to a triangle that was hit. 
From the IDs of those vertices, the 3D coordinates from the vertex positions are extracted in the mesh.
Finally, the exact point needs to be weighted by the vertices of its triangle because only the relative positions from the vertices are known \cite{xia2022bary}. Those relative positions are named u, v, s. 
The results only return two barycentric coordinates for each hit point since the sum of the three coordinates equals one:
\begin{equation} \label{eq:barycentricCoords}
1 = u + v + s
\end{equation}

This means that $s$ can be directly derived from $u$ and $v$.
The weighting to receive the final coordinates works as follows:
\begin{equation} \label{eq:Points}
P = u*P_{vertex_{1}} + v*P_{vertex_{2}} + s*P_{vertex_{3}}
\end{equation}

\subsection{Calculating the Trajectory}

The vehicle's trajectory is simultaneously calculated by driving and reading the generated images. To this end, we leverage the spatial resection concept. Spatial resection is a set of statistical formulas to compute the accuracy of observations by finding a model that fits them best. Consequently, the accuracy of the observations and related values can be calculated afterward. In this case, the camera position is estimated while the observations are optimized at the same time. The position calculation is performed with the help of the photogrammetrical co-linearity equations, which can be seen in \autoref{eq:colin1} and \autoref{eq:colin2}.
{\setlength{\mathindent}{0cm}
\begin{equation} \label{eq:colin1}
\noindent x = \hat{x_0} + z \frac{(\hat{r_{11}}(X-\hat{X_0}) + \hat{r_{21}}(Y-\hat{Y_0}) + \hat{r_{31}}(Z-\hat{Z_0})} {(\hat{r_{13}}(X-\hat{X_0}) + \hat{r_{23}}(Y-\hat{Y_0}) + \hat{r_{33}}(Z-\hat{Z_0})}
\end{equation}
\begin{equation} \label{eq:colin2}
y = \hat{y_0} + z \frac{(\hat{r_{12}}(X-\hat{X_0}) + \hat{r_{22}}(Y-\hat{Y_0}) + \hat{r_{32}}(Z-\hat{Z_0})} {(\hat{r_{13}}(X-\hat{X_0}) + \hat{r_{23}}(Y-\hat{Y_0}) + \hat{r_{33}}(Z-\hat{Z_0})}
\end{equation}}

For the calculation, six values are to be determined, representing the camera's position ($X_0, Y_0, Z_0$) and orientation ($\omega, \phi, \kappa$).
Consequently, for the solution, at least six corresponding points are required. More corresponding points are preferred to get even better accuracy. 

The spatial resection is performed as follows. A direct linear transformation generates the approximate camera orientation values. Input parameters are at least six corresponding points. The approximate camera position is taken from the corresponding GNSS point. The approximate values are required because the spatial resection needs starting points to optimize. Otherwise, the algorithm might run into a local minimum and keeps iterating in an infinite loop. 

In the optimization loop, the derivations of the A-matrix are computed. This is necessary because the refinements of the observations for the best-fitting model are calculated with the help of this matrix. The A-matrix has the size of [n x m], while n is the number of observations and m is the number of values to be determined.

The values that are to be determined are estimated in every iteration by calculating the following:
\begin{equation} \label{eq:delta_x_hat}
\delta \hat x = \frac{(A^T*P_{bb}*A)}{(A^T*P_{bb}*w)}
\end{equation}
Where A is the already described A-matrix. $P_{bb}$ is a matrix that contains weights. Those weights decide how important an observation is, e.g., if the measurement is probably more precise than others because of the geometrical position. The $w$-vector contains the current errors between the approximated $\delta \hat x$ and the actual observations. The optimized observation values are computed by:
\begin{equation} \label{eq:delta_v_hat}
\delta \hat v = A*\delta \hat x - w
\end{equation}

\section{Experiments} \label{sec:experiments}

\begin{figure}[h!]
    \centering
    \includegraphics[width=3.2in]{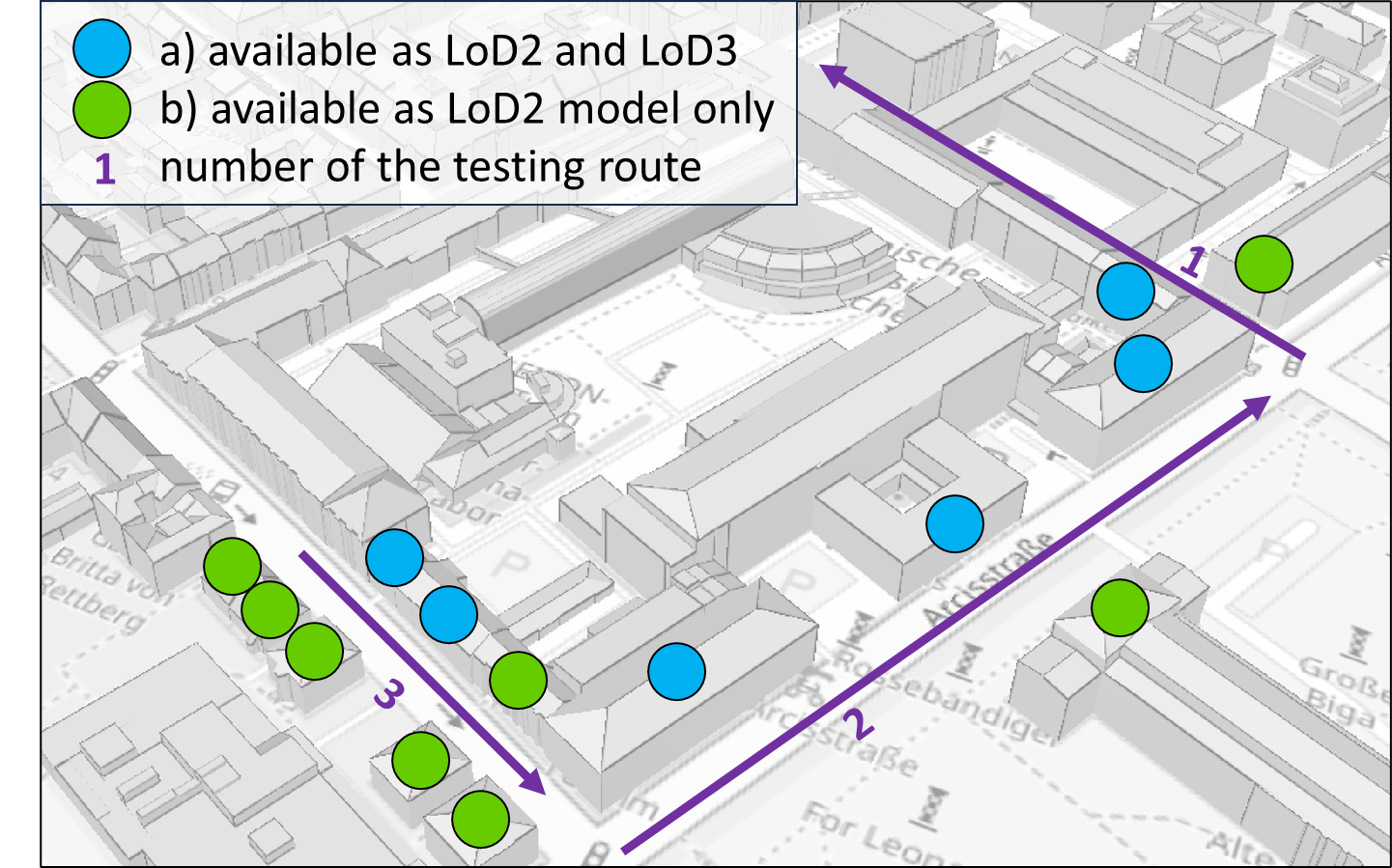}
    \caption{Visualization of the test area divided into three sub-areas (purple) with different availability of building models: a) both in LoD2 and LoD3 (blue), and b) solely in LoD2 (green).} 
    \label{fig:TestingArea}
\end{figure}

The developed method is evaluated in three different test areas, which are visualized in Figure \ref{fig:TestingArea}. 

Notably, there were no instances in the test area where LoD3 models were present on both sides of the street, which mimics a frequently present real-world situation. 
The missing LoD3 models were supplemented from the LoD2 dataset as a solution. 
Therefore, the same models from the LoD2 dataset were used for experiments involving solely LoD2 models. 
This hybrid approach ensures a balanced distribution of potential reference points and promotes an effective configuration. 
Owing to the experimental setup, we were able to directly evaluate the influence of the LoD3 over to the usage of only LoD2 models.

The test site was located in Munich, Germany, and represented a Central European architecture comprising a mixed residential area and a university campus.
The images were acquired by a mobile mapping vehicle of 3D Mapping Solutions GmbH \cite{3DMapping}.
The LoD2 models were downloaded through the governmental open-geodata portal of Bavaria, Germany \cite{testingArea}, whereas the LoD3 models were developed within the interdisciplinary tun2twin project \cite{tum2twin}.

\subsection{Different Approaches for Feature Matching}

The feature matching was performed on five different pillars. Each approach is described in the following subsections.

\subsubsection{Corresponding Image-Pairs of Different Types}
\hfill\\ \vspace{-10pt}

This approach aims to find features from image pairs that are not the same type. This means that one virtual image was matched with one image of reality. Both images were taken from the same GNSS point. The results show only a median of 9-18 suitable feature matches for each image pair. 

\subsubsection{Segmented Images with Deep Learning} \label{subsec:segmentation}
\hfill\\ \vspace{-10pt}

Matching real images with virtual images presents a challenge due to disruptive elements not being shared between the two image types. To mitigate this issue, the approach involves segmenting the real images. 
The used algorithm was built on the SegFormer network \cite{xie2021segformer}, using the MMSegmentation framework \cite{mmsegmentation}. 
This algorithm generated an output image where distinct elements (e.g., buildings) were each assigned a unique color.

To match these semantically segmented images, a method is devised wherein the segmented images act as masks for the real images to isolate the building areas from the real images. 
Consequently, only the building-related parts of the real image are retained and matched with their virtual counterparts. 
This strategic approach effectively minimizes any potential disruptions from extraneous objects that could potentially result in erroneous matches.

\subsubsection{Sobel-Filter and Canny Edge Detector}
\hfill\\ \vspace{-10pt}

An established approaches for feature matching focuse solely on features along object edges, in this case, including structures such as window edges, e.g., Sobel filters and Canny edge detector \cite{singh2015comparison}. 

In instances where it was desirable to exclusively match buildings, applying a bounding box to actual images proved effective. 
These filters were harmonized with the segmented images detailed in Section \ref{subsec:segmentation} to ensure precision. 
This strategic combination ensured that only relevant objects for vehicle localization were considered. 

\section{Results and Discussion}

The experiments were evaluated by comparing the number of found features and the standard deviation of the camera position. 
For every table, the corresponding gain was computed as a percentage increase or decrease between LoD2 and LoD3. 

\subsection{Number of found features}
\begin{figure}[h!]
    \centering
    \includegraphics[width=3.1in]{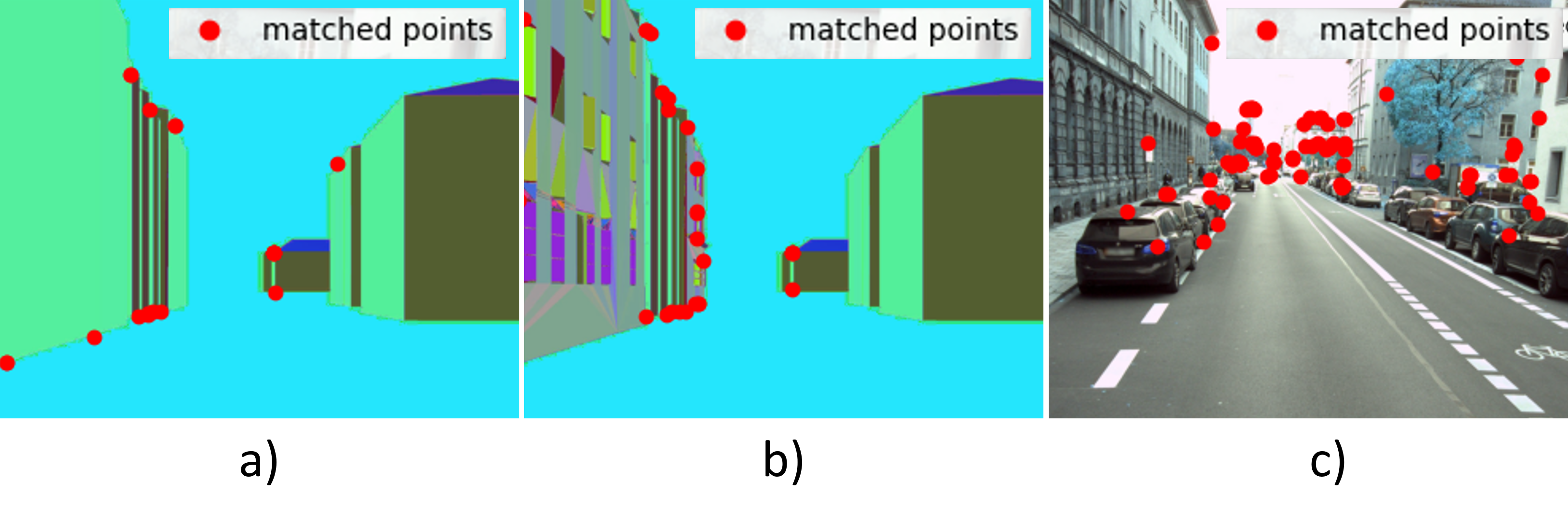}
    \caption{Matched features a) LoD2 b) LoD3 c) optical image.}
    \label{fig:FeatComp}
\end{figure}

The different approaches led to different results regarding the number of found features, which represented those that can be successfully matched between two images. 
The selection pertains to the median of all tested images within each area. The best result for each experiment is written in bold, comparing LoD2 and LoD3. An example of the found features in the images is visualized in Figure \ref{fig:FeatComp}.

A comprehensive comparison for each model, namely LoD2 and LoD3, is presented across different test areas in Tables \ref{tab:comparisonFeatures1}, \ref{tab:comparisonFeatures2}, and \ref{tab:comparisonFeatures3}.

\begin{table}[h!]
    \caption{Number of Found Features (Median) and Gain in the Test Area 1.}
    \centering
    \begin{tabular}{c|c|c|c}
      & LoD2 & LoD3 & gained\\
    \hline
    Corresponding images & 17  & \textbf{18} & 6\% \\
    Feature images & 20 & \textbf{50}  & 60\%\\
    Sobel-filter & \textbf{14}  & 13 & -8\%\\
    Canny edge detection & 21  & 21 & 0\%\\
    Mask & 18  & 18 & 0\%\\
    Mask and Sobel-filter & \textbf{21}  & 11  & -91\%\\
    Mask and Canny edge detection & \textbf{34}  & 22 & -55\%\\
    \hline
    \end{tabular}
    \label{tab:comparisonFeatures1}
\end{table}

\begin{table}[h!]
    \caption{Number of Found Features (Median) and Gain in the Test Area 2.}
    \centering
    \begin{tabular}{c|c|c|c}
      & LoD2 & LoD3 & gained\\
    \hline
    Corresponding images & \textbf{12}  & 9 & -33\%\\
    Feature images & 20  & \textbf{65}  & 69\%\\
    Sobel-filter & 10  & 10 & 0\% \\
    Canny edge detection & 8  & \textbf{14} & 43\%\\
    Mask & \textbf{15}  & 12 & -25\%\\
    Mask and Sobel-filter & 13  & \textbf{14}  & 7\%\\
    Mask and Canny edge detection & 21  & \textbf{27} & 22\% \\
    \hline
    \end{tabular}
    \label{tab:comparisonFeatures2}
\end{table}

\begin{table}[h!]
    \caption{Number of Found Features (Median) and Gain in the Test Area 3.}
    \centering
    \begin{tabular}{c|c|c|c}
      & LoD2 & LoD3 & gained\\
    \hline
    Corresponding images & 15 & \textbf{18} & 17\%\\
    Feature images & 20 & \textbf{47} & 57\%\\
    Sobel-filter & 13 & \textbf{15} & 13\%\\
    Canny edge detection & 14 & \textbf{20} & 30\%\\
    Mask & 13 & \textbf{20} & 35\%\\
    Mask and Sobel-filter & 15 & \textbf{17} & 12\%\\
    Mask and Canny edge detection & \textbf{26} & 21 & -24\%\\
    \hline
    \end{tabular}
    \label{tab:comparisonFeatures3}
\end{table}

As our experiments corroborate, the LoD3 models, on average, result in the same or more found features between the images. We deem the feature images algorithm as the most reliable feature extraction between semantic 3D building models and optical images, as it scored more features on average in the three test areas (60\%, 69\%, and 58\%, respectively). 

\begin{figure}[h!]
    \centering
    \includegraphics[width=3.1in]{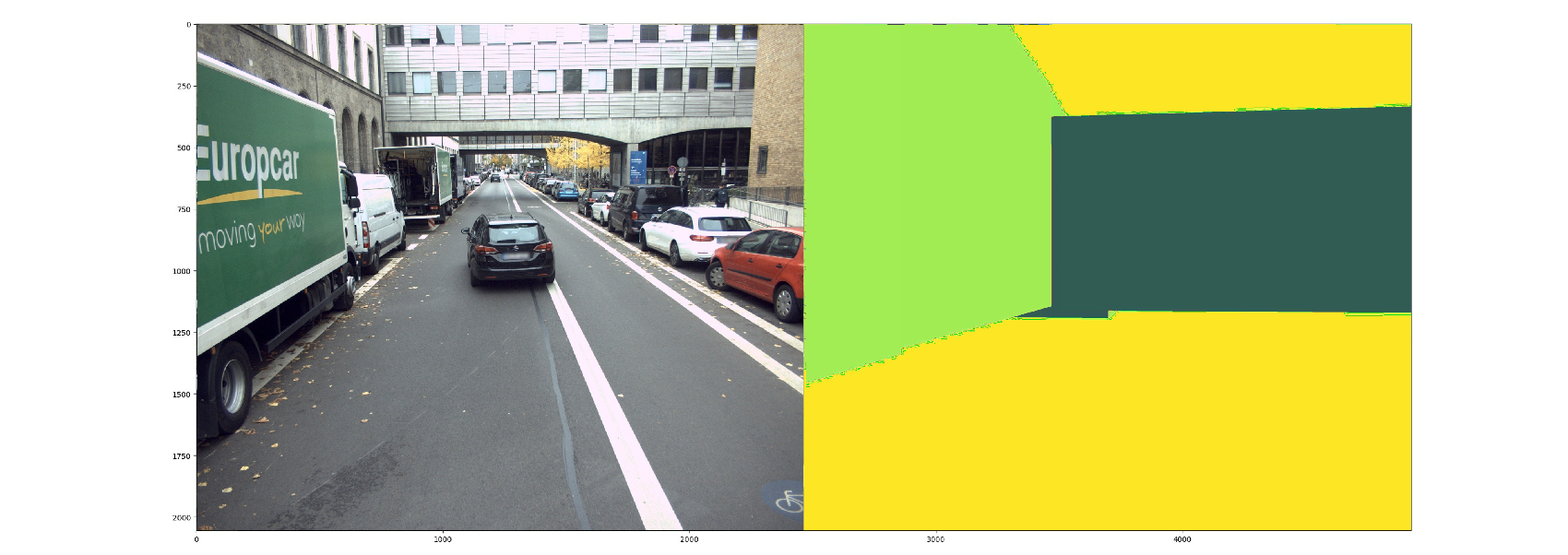}
    \caption{Visualization of an absent underpass in LoD2.}
    \label{fig:underpass}
\end{figure}

In the test area 1 with prevalent LoD2 models, the algorithm matched features for LoD2 until the point where the whole virtual image was facing the model-absent underpass. 
Then, no feature matching was possible again and the algorithm failed.
This happens due to the frequently absent underpasses in LoD2, where instead a vertical wall is rendered to the ground level, as seen in Figure \ref{fig:underpass}.

Especially in test area 2, more features lead to a more reliable estimation of the trajectory point: up to 69\% more features (see Table \ref{tab:comparisonFeatures2}). 

As described in Section \ref{sec:experiments}, test area 3 exhibited the optimal situation for localizing the vehicle with the help of high-detail polyhedral models. 
This can be seen in Table \ref{tab:comparisonFeatures3}, where 57\% more features were found, and seven of the tested methods achieved an increase of found features.
We also observe that the higher the buildings, the more details can be found for feature matching.
Furthermore, narrow streets are advantageous because the buildings are closer to the camera and can be captured in more detail. 
In contrast, GNSS data is most accurate in urban environments when surrounded by only low buildings; narrow streets lead to positionining deviations. 
Consequently, the localization with images and LoD3 models can be advantageous over using GNSS in urban areas. 
Both methods are like counterparts: where one method fails, the other has advantages.
\subsection{Positioning accuracy}
To analyze the advantage can be achieved by LoD3 models, Tables \ref{tab:comparisonStd1}, \ref{tab:comparisonStd2}, \ref{tab:comparisonStd3} show the accuracy of the estimated camera position in comparison.

\begin{table}[h!]
    \caption{Standard Deviation of the Camera Position (Median) and Gain in Test Area 1.}
    \centering
    \begin{tabular}{c c|c|c|c}
     & & LoD2 & LoD3 & gained\\
    \hline
    Corresponding images & \makecell{$\sigma_X$ \\ $\sigma_Y$ \\ $\sigma_Z$}& \makecell{{35.81m} \\ \textbf{36.39m} \\ 44.25m} & \makecell{\textbf{15.26m} \\ 46.10m \\ \textbf{21.11m}} & \makecell{57\%  \\-27\% \\ 52\%}\\
    \hline
    Feature images & \makecell{$\sigma_X$ \\ $\sigma_Y$ \\ $\sigma_Z$}& \makecell{\textbf{81.72m} \\ \textbf{50.40m} \\ \textbf{74.92m}} & \makecell{111.87m \\ 152.39m \\ 94.45m}  & \makecell{-37\%  \\-202\% \\ -26\%}\\
    \hline
    Sobel-filter & \makecell{$\sigma_X$ \\ $\sigma_Y$ \\ $\sigma_Z$} & \makecell{66.13m \\ 67.34m \\ 65.85m} & \makecell{\textbf{51.23m} \\ \textbf{58.56m} \\ \textbf{31.64m}} & \makecell{23\%  \\13\% \\ 52\%}\\
    \hline
    Canny edge detection & \makecell{$\sigma_X$ \\ $\sigma_Y$ \\ $\sigma_Z$}& \makecell{48.71m \\ 50.27m \\ 52.06m} & \makecell{\textbf{35.82m} \\ \textbf{47.18m} \\ \textbf{27.31m}} & \makecell{26\%  \\6\% \\ 48\%}\\
    \hline
    Mask & \makecell{$\sigma_X$ \\ $\sigma_Y$ \\ $\sigma_Z$}& \makecell{\textbf{19.93m} \\ \textbf{32.32m} \\ \textbf{29.15m}} & \makecell{56.26m \\ 44.57m \\ 48.74m} & \makecell{-182\%  \\-38\% \\ -67\%}\\
    \hline
    Mask and sobel-filter & \makecell{$\sigma_X$ \\ $\sigma_Y$ \\ $\sigma_Z$}& \makecell{\textbf{55.96m} \\ 74.12m \\ 59.12m} & \makecell{65.78m \\ \textbf{23.43m} \\ \textbf{42.22m}} & \makecell{-18\%  \\68\% \\ 29\%}\\
    \hline
   \makecell{Mask and canny \\ edge detection}  & \makecell{$\sigma_X$ \\ $\sigma_Y$ \\ $\sigma_Z$} & \makecell{59.45m \\ 92.93m \\ 50.31m} & \makecell{\textbf{38.27m} \\ \textbf{55.16m} \\ \textbf{42.67m}} & \makecell{36\%  \\41\% \\ 15\%}\\
    \end{tabular}
    \label{tab:comparisonStd1}
\end{table}

\begin{table}[h!]
    \caption{Standard Deviation of the Camera Position (Median) and Gain in Test Area 2.}
    \centering
    \begin{tabular}{c c|c|c|c}
     & & LoD2 & LoD3 & gained\\
    \hline
    Corresponding images & \makecell{$\sigma_X$ \\ $\sigma_Y$ \\ $\sigma_Z$}& \makecell{48.02m \\ 44.22m \\ 58.98m} & \makecell{\textbf{17.49m} \\ \textbf{41.67m} \\ \textbf{49.33m}} & \makecell{64\%  \\6\% \\ 16\%}\\
    \hline
    Feature images & \makecell{$\sigma_X$ \\ $\sigma_Y$ \\ $\sigma_Z$}& \makecell{\textbf{130.94m} \\ 191.27m \\ 194.87m} & \makecell{168.78m \\ \textbf{169.88m} \\ \textbf{158.22m}} & \makecell{-29\%  \\11\% \\ 19\%}\\
    \hline
    Sobel-filter & \makecell{$\sigma_X$ \\ $\sigma_Y$ \\ $\sigma_Z$}& \makecell{93.47m \\ 76.02m \\ 51.87m} & \makecell{\textbf{21.96m} \\ \textbf{26.54m} \\ \textbf{15.89m}} & \makecell{77\%  \\65\% \\ 69\%}\\
    \hline
    Canny edge detection & \makecell{$\sigma_X$ \\ $\sigma_Y$ \\ $\sigma_Z$}& \makecell{\textbf{43.62m} \\ \textbf{24.39m} \\ 38.84m} & \makecell{48.33m \\ 52.87m \\ \textbf{33.06m}} & \makecell{-11\%  \\-117\% \\ 15\%}\\
    \hline
    Mask & \makecell{$\sigma_X$ \\ $\sigma_Y$ \\ $\sigma_Z$}& \makecell{\textbf{9.62m} \\ \textbf{13.59m} \\ 14.63m} & \makecell{13.77m \\ 25.69m \\ \textbf{10.12m}} & \makecell{-43\%  \\-89\% \\ 31\%}\\
    \hline
    Mask and sobel-filter & \makecell{$\sigma_X$ \\ $\sigma_Y$ \\ $\sigma_Z$}& \makecell{\textbf{67.45m} \\ 91.99m \\ 96.01m} & \makecell{94.36m \\ \textbf{85.56m} \\ \textbf{80.71m}} & \makecell{-40\%  \\7\% \\ 16\%}\\
    \hline
    \makecell{Mask and canny \\ edge detection}  & \makecell{$\sigma_X$ \\ $\sigma_Y$ \\ $\sigma_Z$} & \makecell{\textbf{37.74m} \\ \textbf{40.91m} \\ 40.69m} & \makecell{42.08m \\ 58.02m \\ \textbf{27.19m}} & \makecell{-11\%  \\-42\% \\ 33\%}\\
    \end{tabular}
    \label{tab:comparisonStd2}
\end{table}

\begin{table}[h!]
    \caption{Standard Deviation of the Camera Position (Median) and Gain in Test Area 3.}
    \centering
    \begin{tabular}{c c|c|c|c}
     & & LoD2 & LoD3 & gained\\
    \hline
    Corresponding images & \makecell{$\sigma_X$ \\ $\sigma_Y$ \\ $\sigma_Z$}& \makecell{71.37m \\ 76.23m \\ 80.26m} & \makecell{\textbf{40.05m} \\ \textbf{66.23m} \\ \textbf{77.21m}} & \makecell{44\%  \\13\% \\ 4\%}\\
    \hline
    Feature images & \makecell{$\sigma_X$ \\ $\sigma_Y$ \\ $\sigma_Z$}& \makecell{171.01m \\ 127.70m \\ 104.11m} & \makecell{\textbf{21.14m} \\ \textbf{38.17m} \\ \textbf{21.44m}} & \makecell{88\%  \\70\%  \\79\%}\\
    \hline
    Sobel-filter & \makecell{$\sigma_X$ \\ $\sigma_Y$ \\ $\sigma_Z$} & \makecell{59.78m \\ \textbf{43.54m} \\ 42.67m} & \makecell{\textbf{40.71m} \\ 65.81m \\ \textbf{35.07m}} & \makecell{32\%  \\-51\%  \\18\%}\\
    \hline
    Canny edge detection & \makecell{$\sigma_X$ \\ $\sigma_Y$ \\ $\sigma_Z$}& \makecell{24.98m \\ 33.31m \\ 19.12m} & \makecell{\textbf{9.35m} \\ \textbf{11.62m} \\ \textbf{6.07m}} & \makecell{63\%  \\65\%  \\68\%}\\
    \hline
    Mask & \makecell{$\sigma_X$ \\ $\sigma_Y$ \\ $\sigma_Z$}& \makecell{38.90m \\ \textbf{33.51m} \\ 40.14m} & \makecell{\textbf{20.10m} \\ 60.34m \\ \textbf{27.97m}} & \makecell{48\%  \\-80\%  \\30\%}\\
    \hline
    Mask and sobel-filter & \makecell{$\sigma_X$ \\ $\sigma_Y$ \\ $\sigma_Z$}& \makecell{\textbf{21.05m} \\ \textbf{23.25m} \\ \textbf{19.21m}} & \makecell{33.62m \\ 47.88m \\ 44.98m} & \makecell{-60\%  \\-106\%  \\-134\%}\\
    \hline
    \makecell{Mask and canny \\ edge detection}  & \makecell{$\sigma_X$ \\ $\sigma_Y$ \\ $\sigma_Z$} & \makecell{159.73m \\ 168.19m \\ 180.55m} & \makecell{\textbf{7.83m} \\ \textbf{13.24m} \\ \textbf{5.19m}} & \makecell{95\%  \\92\%  \\97\%}\\
    \end{tabular}
    \label{tab:comparisonStd3}
\end{table}

Regarding test area 1, the accuracy of the feature images algorithm drops to [-37\%, -202\%, -26\%]. This shows, that the algorithm is not yet working perfectly for every situation. The configuration with the underpass leads also in LoD3 to problems in the accuracies. Still, other methods show the advantages of LoD3 also in this case. The tackling of the problems is described in Section \ref{sec:discussion}.

The developed algorithm works less accurately in urban areas when buildings are available on only one side of the street, see Tables \ref{tab:comparisonFeatures2} and \ref{tab:comparisonStd2} (test area 2), especially for the Feature images algorithm. 

As seen in Tables \ref{tab:comparisonFeatures1}, \ref{tab:comparisonFeatures2} and \ref{tab:comparisonFeatures3}, LoD3 reaches a higher number of features. More found features provide a better distribution in the image and, therefore, result in a better standard deviation of the camera point, e.g., up to [88\%, 70\%, 79\%] more accurate in test area 3 (see Table \ref{tab:comparisonFeatures3}).  
Considering the standard deviation of the camera position for the evaluation, LoD3 is favorable in test area 3.

\subsection{Limitations} \label{sec:discussion}
The developed algorithm shows the advantages of using LoD3 over LoD2 models. Still, the deviations of the estimated point from the corresponding GNSS point are large. 
To use those models for positioning, the feature matching has to be optimized so that matched features refer to exactly one single point in the real world. 
This is not the case for all matched features so far.

In general, the deviation of the calculated camera position performs slightly better in LoD3 compared to LoD2. 
A reason for the deviation from the GNSS point is an imprecise calculation of the 3D coordinates. 
The exact coordinates are weighted by the barycentric coordinates in the triangle. 
Consequently, each triangle vertex belongs to exactly one of the barycentric coordinates $u, v, s$. 
This depends on the sequence in which the triangle coordinates are saved in the mesh. The order of the vertices is a requirement currently investigated for LoD2 and LoD3 models. 
The topology for such models has to be consistent in every model, e.g., in the right-hand rule. 
This is not yet common for all developed LoD3 models since the automatic generation remains in its infancy and is prone to manual, randomly induced errors. 
These errors lead to an imprecise calculation of the 3D coordinates and, therefore, to a deviation of the calculated camera position from the GNSS point. 
The method developed in this paper works assuming that the models are correct and the vertices are saved in the right-hand rule. 

\section{Conclusion}
In this paper, we propose an approach to compare the performance of LoD2 against LoD3 models for map- and image-based vehicle positioning. 
Using a feature image method, we harmonize semantic 3D city models with optical images. 
Our experiments corroborate that although we had different representations, optical images vs. quasi-optical images of the LoD models, our method performs well in feature matching. 
The presented method performs best when buildings cover both sides of the street.
We also observe that the higher the building models, the more details can be found for feature matching.
Additionally, narrow streets are advantageous because the buildings are closer to the camera and can be captured in more detail. 
We are convinced our image and model-based navigation algorithm can inspire upcoming work. 
In the future, we plan to improve upon measurements, e.g., by using SLAM. With this, we hope to decrease the large deviations in our calculations from the GNSS points. Additionally, we will work on reducing false feature matches to achieve higher accuracy. 
We also plan to do more experiments on more datasets and aim to expand the LoD3 datasets.

\section*{Acknowledgment}
The authors would like to thank 3D Mapping Solutions GmbH for providing the images to develop and test the algorithm.
This work was supported by the Bavarian State Ministry for Economic Affairs, Regional Development and Energy within the framework of the IuK Bayern project \textit{MoFa3D - Mobile Erfassung von Fassaden mittels 3D Punktwolken}, Grant No.\ IUK643/001.
Moreover, the work was conducted within the framework of the {Leonhard Obermeyer Center} at the Technical University of Munich (TUM).
We gratefully acknowledge the tum2twin team at the TUM for the valuable insights and for providing the CityGML datasets.


{
	\begin{spacing}{1.17}
		\normalsize
		\bibliography{bibliography} 
	\end{spacing}
}

\end{document}